\documentclass[conference]{IEEEtran}
\ifCLASSINFOpdf
\else
\fi
\usepackage{multirow}
\usepackage{booktabs}
\usepackage{graphics}
\usepackage{adjustbox}
\usepackage{subcaption}
\usepackage[font=small,labelfont=bf]{caption}
\usepackage{url}
\usepackage{graphicx}
\usepackage{array}

\begin{document}
%
\title{Impact of Video Processing Operations in Deepfake Detection}

\author{\IEEEauthorblockN{Yuhang Lu}
\IEEEauthorblockA{
Multimedia Signal Processing Group (MMSPG) \\
\'Ecole Polytechnique F\'ed\'erale de Lausanne (EPFL)\\
Lausanne, Switzerland\\
Email: yuhang.lu@epfl.ch}
\and
\IEEEauthorblockN{Touradj Ebrahimi}
\IEEEauthorblockA{
Multimedia Signal Processing Group (MMSPG) \\
\'Ecole Polytechnique F\'ed\'erale de Lausanne (EPFL)\\
Lausanne, Switzerland\\
Email: touradj.ebrahimi@epfl.ch}
}

%


\maketitle

\begin{abstract}
The detection of digital face manipulation in video has attracted extensive attention due to the increased risk to public trust. To counteract the malicious usage of such techniques, deep learning-based deepfake detection methods have been developed and have shown impressive results. However, the performance of these detectors is often evaluated using benchmarks that hardly reflect real-world situations. For example, the impact of various video processing operations on detection accuracy has not been systematically assessed. To address this gap, this paper first analyzes numerous real-world influencing factors and typical video processing operations. Then, a more systematic assessment methodology is proposed, which allows for a quantitative evaluation of a detector's robustness under the influence of different processing operations. Moreover, substantial experiments have been carried out on three popular deepfake detectors, which give detailed analyses on the impact of each operation and bring insights to foster future research.


\end{abstract}


%
\IEEEpeerreviewmaketitle

\section{Introduction}

Recent years have witnessed remarkable progress in computer vision tasks due to the rapid development of deep convolutional neural networks (DCNNs) and the ease of obtaining large-scale datasets. This advancement has also led to an increase in new applications. For instance, generative adversarial networks (GANs) \cite{karras2017progressive, karras2019style, karras2020analyzing} have made it possible to produce fake content that appears authentic to human eyes. In fact, deep learning-based face manipulation techniques \cite{roessler2019faceforensicspp, thies2019deferred, nirkin2019fsgan,zakharov2019few} can alter the expression, attributes, and even identity of a human face image, the outcome of which is referred to by the term `Deepfake'. The development of such technologies and the wide availability of open-source software have simplified the creation of deepfakes, causing significant public concern and undermining our trust in online media. As a result, detecting manipulations in facial video has become a popular topic in the media forensics community, receiving increased attention from academia and businesses to counteract the misuse of these deepfake techniques and malicious attacks.

Nowadays, with the aid of advanced deep-learning techniques and various large-scale datasets, numerous detection methods \cite{roessler2019faceforensicspp, Nguyen2019UseOA, Zhao2021MultiattentionalDD, liu2021spatial, qian2020thinking, li2021frequency, luo2021generalizing} have been published and have shown remarkable results. But  recent studies \cite{khodabakhsh2018fake,xuan2019generalization} have shown that the detection accuracy significantly drops in the cross-dataset scenario, where the fake samples are created by unknown manipulation methods. As a result, cross-dataset evaluation has become an important step to better demonstrate the generalization ability of different types of detectors. 

Nevertheless, researchers still overlook a common realistic scenario where DCNN-based methods are susceptible to real-world perturbations and processing operations. 
In more realistic conditions, video content on social media can encounter unpredictable distortions during propagation due to extrinsic or artificial impacts, such as varying brightness, compression, and video filters, to mention a few. 
In practice, most of the current deep learning-based deepfake detectors are developed and evaluated using constrained and high-quality face manipulation datasets and benchmarks, making them insufficiently robust to real-world situations. Our previous study \cite{lu2022novel} has contributed with a realistic evaluation framework for deepfake detectors with a specific focus on images. However, a large number of deepfakes are distributed in the form of video on social media. Therefore, this paper investigates potential influencing factors for deepfake detectors in more practical scenarios and proposes a reliable method for evaluation. Ideally, this study will bring valuable insights and encourage researchers to create more resilient detection techniques. 
In summary, this paper makes the following contributions. 


\begin{itemize}
    \item It analyzes numerous video processing operations that can potentially impact the performance of a deepfake detector. 
    \item A systematic evaluation method is proposed to evaluate the performance of learning-based deepfake detection systems affected by different influencing factors.
    \item It presents substantial experimental results to measure and analyze the performance of three popular deepfake detection techniques.
    
\end{itemize}

\section{Related Work}
\subsection{Deepfake Detection}
In computer vision, detecting deepfakes is typically approached as a binary classification. Initially, methods that relied on facial expressions \cite{Agarwal_2019_CVPR_Workshops}, head movements \cite{Yang2019ExposingDF}, and eye blinking \cite{9072088} were suggested as solutions to tackle this detection challenge. In recent years, the main approach to address this problem is to utilize deep learning technology with sophisticated neural networks.
\cite{zhou_two-stream_2017} first proposed a two-stream neural network to detect deepfakes. Nguyen et al. \cite{Nguyen2019UseOA} employed a combination of conventional CNNs and Capsule networks and surpassed the benchmark at that time. Rössler et al. \cite{roessler2019faceforensicspp} adopted the well-known XceptionNet as one of the baseline models alongside three other learning-based methods to perform deepfake detection. 
Video deepfake detectors \cite{guera2018deepfake, sabir2019recurrent} leverage recurrent neural networks (RNNs) to capture forgery traces from temporal sequences. Several innovative network architectures, such as efficient networks \cite{Montserrat_2020_CVPR_Workshops, Shiohara_2022_CVPR}, and vision transformers \cite{zheng2021exploring}, have also been explored for deepfake detection tasks. 
In \cite{dang2020detection}, the well-known attention mechanism was applied to the deepfake detection system, which highlighted the informative regions. 
Other attempts have been made by \cite{liu2021spatial,qian2020thinking,li2021frequency} to analyze forgery contents in the frequency domain. In general, these methods separate the information contained in images according to the frequency bands via FFT or DCT transformation and more effectively capture traces of forgery. 
Another branch of research directly works on authentic video and generates synthetic faces during a training process. 
For example, Xray \cite{Li2019OnLF} and SBIs \cite{Shiohara_2022_CVPR} methods reproduce the blending artifacts that exist in many types of deepfakes and force the network to learn more generic representations. These methods show a strong generalization ability but remain susceptible to common real-world perturbations.

\subsection{Deepfake Benchmarks}

To facilitate a faster progress and better performance in deepfake detection, numerous benchmarks, competitions, and challenges have been organized by academia and businesses. 
Early on, UADFV \cite{yang2019exposing} was one of the first public databases, comprising 49 real and fake video. Korshunov and Marcel contributed a larger database in \cite{korshunov2018deepfakes}, called the Deepfake-TIMIT, which comprises 620 fake video of 32 subjects created by GAN-based face-swapping algorithms. 
Celeb-DF \cite{Celeb_DF_cvpr20} provides a high-quality face forgery detection benchmark and often used for cross-dataset evaluation.
One of the most popular databases and benchmarks is FaceForensics++ \cite{roessler2019faceforensicspp}. It contains a larger number of fake video footage generated by different tools and presents six baseline detectors for comparison. It is important to remark that FaceForensics++ first considers the different levels of video quality and simulates the video compression techniques applied in social media. But a deeper exploration of more perturbation types is expected. Later on, Jiang et al. \cite{jiang2020deeperforensics10} presented a large-scale benchmark for face forgery detection, called DeeperForensics-1.0. Facebook also launched the Deepfake Detection Challenge (DFDC) \cite{DFDC2020} in collaboration with other companies and institutions. Although both benchmarks comprise a hidden test set impacted by realistic perturbations, their assessment approach either ignores many practical influencing factors or employs a less reliable evaluation strategy. \cite{lu2022novel} introduced a more rigorous way to measure the influence of image processing operations. This paper adopts a similar evaluation method and when compared to previous work, pays special attention to video-processing operations that exist in the real world.




\section{Proposed Method}

\subsection{Realistic Influencing Factors for Video Deepfakes}
\newcommand\w{0.25\linewidth}
\newcommand\y{\linewidth}
\begin{figure}[t]
\centering
\begin{subfigure}[b]{\w}
  \includegraphics[width=\y]{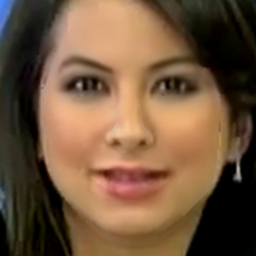}  
  \caption{Unaltered}
\end{subfigure}%
\hfill
\begin{subfigure}[b]{\w}
  \includegraphics[width=\y]{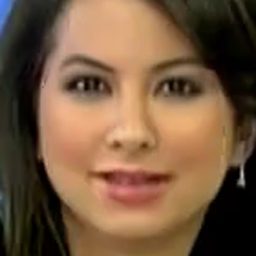}  
  \caption{C23}
\end{subfigure}%
\hfill
\begin{subfigure}[b]{\w}
  \includegraphics[width=\y]{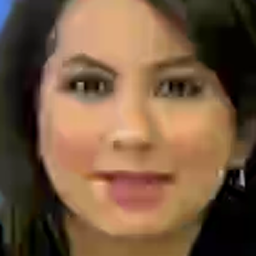}  
  \caption{C40}
\end{subfigure}%
\hfill
\begin{subfigure}[b]{\w}
  \includegraphics[width=\y]{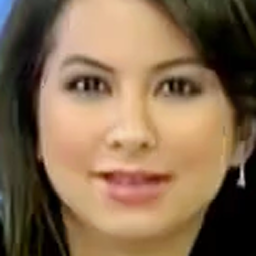}  
  \caption{Light}
\end{subfigure}%
\hfill
\begin{subfigure}[b]{\w}
  \includegraphics[width=\y]{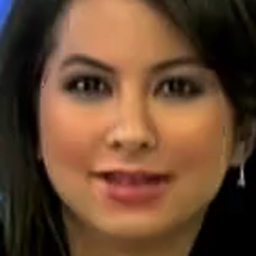}  
  \caption{Dark}
\end{subfigure}%
\hfill
\begin{subfigure}[b]{\w}
  \includegraphics[width=\y]{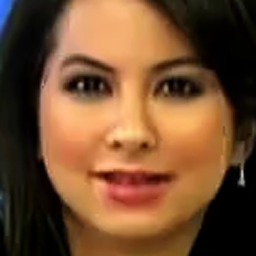}
  \caption{Contrast}
\end{subfigure}%
\hfill
\begin{subfigure}[b]{\w}
  \includegraphics[width=\y]{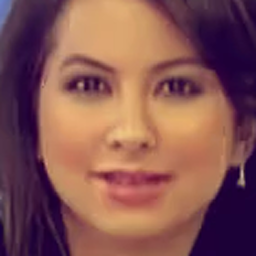}
  \caption{Vintage}
\end{subfigure}%
\hfill
\begin{subfigure}[b]{\w}
  \includegraphics[width=\y]{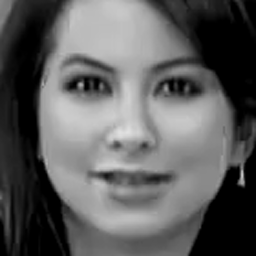}  
  \caption{Grayscale}
\end{subfigure}%
\hfill
\begin{subfigure}[b]{\w}
  \includegraphics[width=\y]{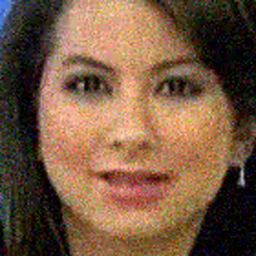}  
  \caption{Noise}
\end{subfigure}%
\hfill
\begin{subfigure}[b]{\w}
  \includegraphics[width=\y]{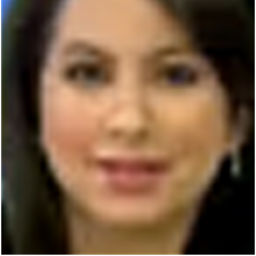}  
  \caption{Resolution}
\end{subfigure}%
\hfill
\begin{subfigure}[b]{\w}
  \includegraphics[width=\y]{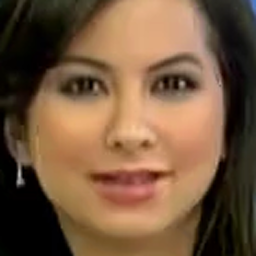} 
  \caption{Hflip}
\end{subfigure}%
\hfill
\begin{subfigure}[b]{\w}
  \includegraphics[width=\y]{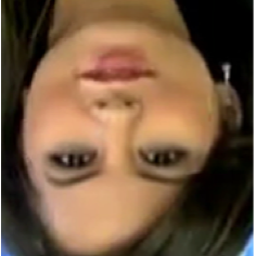}  
  \caption{Vflip}
\end{subfigure}
\caption{Example of a typical video frame in the FFpp test set after applying different video processing operations. Some notations are explained as follows. C23 and C40: Video compression using H.264 codec with factors of 23 and 40. Light and Dark: Increase and decrease brightness. Resolution: Reduce video resolution. Hflip and Vflip: Horizontal and Vertical flip.}
\label{fig:video-op-example}
\end{figure}

\begin{figure*}[t]
	\centering
	\begin{adjustbox}{width=0.75\textwidth}
    \includegraphics[]{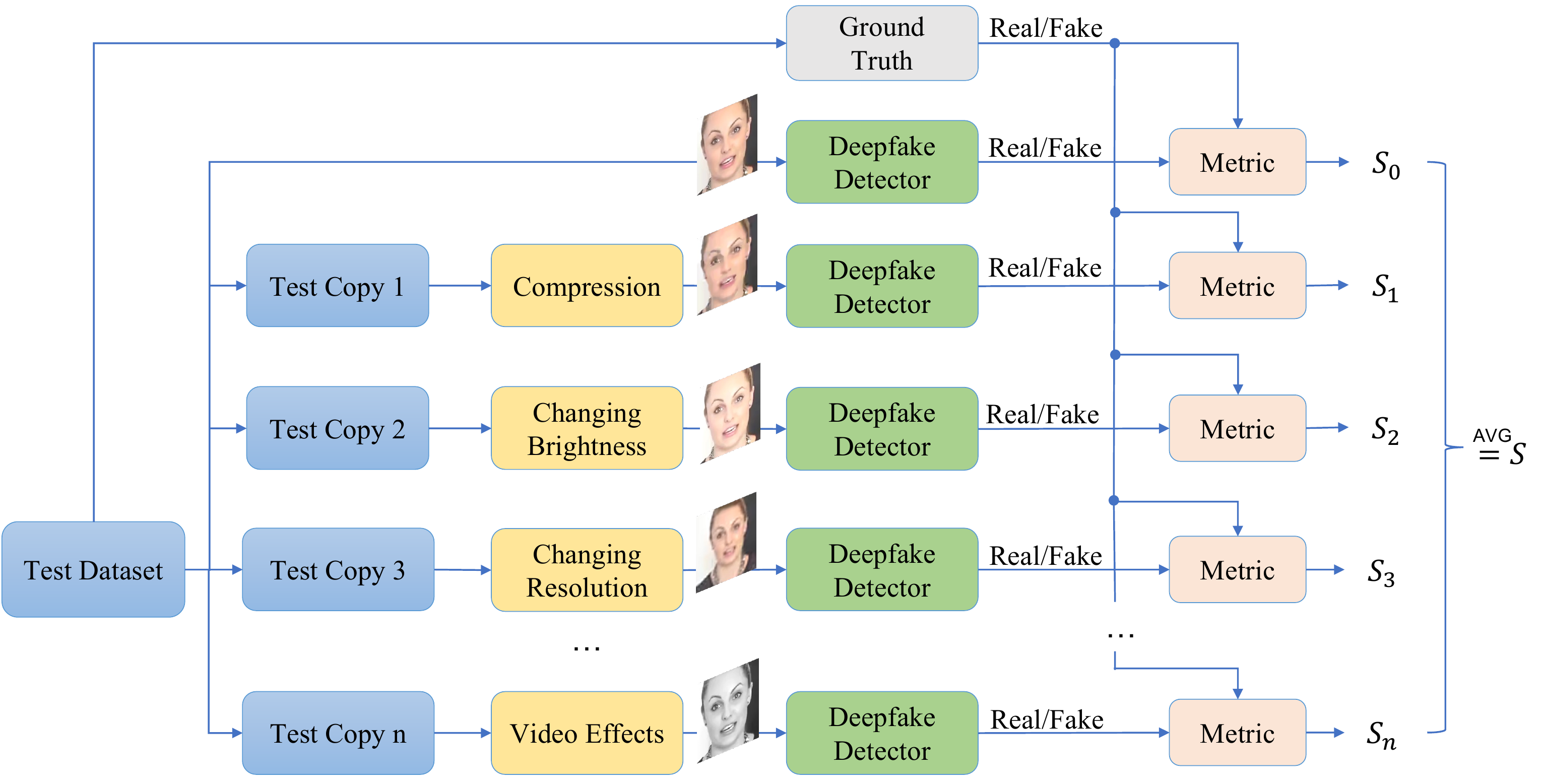}
	\end{adjustbox}
	\caption{Proposed assessment method for measuring the impact of multiple influencing factors.}
	\label{fig:af}
\end{figure*}


The processing operations and various video effects are widespread on different social media, smartphone applications, and streaming platforms. Their impact on the accuracy of detection methods should not be neglected.
This paper studies seven categories of video processing operations with the most commonly used parameters. The factors are also described in detail as follows and the illustrative examples of testing data are shown in Figure \ref{fig:video-op-example}.

\textbf{Video Compression:}
Video compression technology can save storage space and allows for a high-quality video to be distributed worldwide. Although lossless video compression codecs can perform at a compression rate of 5 to 12, lossy compression video can achieve a much lower data rate while maintaining high visual quality. 
The deepfake video can be compressed multiple times before propagating through social networks. However, the side effects of lossy compression artifacts on deep learning-based detectors have not been sufficiently studied. 
It is necessary to evaluate the robustness of a deepfake detector on compressed deepfake video. In this context, the proposed assessment framework includes test data compressed by H.264 codec with two constant rate factors, namely 23 and 40, using the FFMPEG toolbox.

\textbf{Flip:}
Horizontally flipping a video creates a mirrored version of the original footage and is a  video editing technique sometimes used for aesthetic reasons. However, the impact of this operation on deepfake detectors has not been assessed before. 
On the other hand, vertically flipping a video is a simple way to deceive a detector, as most current detection methods do not adjust or correct the face pose during preprocessing. Hence, a flipped video can be uploaded and remain undetected on social networks while still being readable to a human.

\textbf{Video Filters:}
Video filters have become popular on social media in recent years. They are preset effects available in many video editing apps, software, and social media platforms, providing easy access for users to alter the look of a video clip. Some common types of video filters include color filters, beauty filters, stylization filters, etc. 
The overall color palette of a deepfake video can be changed by a video filter on social media, making it an out-of-distribution sample from common deepfake databases. 
In the proposed assessment framework, two common filters, namely `Vintage' and `Grayscale', are taken into consideration.

\textbf{Brightness:}
Brightness refers to the overall lightness or darkness of a video. Modifying the brightness can impact the perception of colors and the visibility of details and textures. For example, increasing brightness can enhance the visibility of details in shadows, while decreasing it can obscure the details. In real-world conditions, the brightness of a video is often adjusted to create a specific style. The proposed assessment framework takes this situation into consideration and measures the performance of a detector under different brightness conditions. More specifically, the `Lighten' and `Darken' commands in the FFMPEG toolbox are applied to the test video respectively.

\textbf{Contrast:}
Contrast refers to the difference between the lightest and darkest areas of a video. Similar to brightness, adjusting contrast is one of the most common operations to change the visual appearance of a video. The framework used in this paper is capable of adjusting the contrast value of the test deepfake video to measure the performance of a detector.

\textbf{Noise:}
Video noise is a common problem in video captured in low-light conditions or using small-sensor devices, such as mobile phones. It often appears as annoying grains and artifacts in the footage. The assessment framework introduces Gaussian noise with a constant intensity but varying temporal distribution to the video data.

\textbf{Resolution:}
Resolution refers to the number of pixels in a video. There is an important trade-off between the resolution and file size. Decreasing the resolution of a video will generally result in a lower-quality video with fewer details to be displayed on the screen, but it can also reduce the file size, which makes it easier to store and share. On the other hand, the resolution change can also affect the ratio of the width to the height of the video. The performance of the deepfake detector when facing low-resolution or stretched video will be evaluated by the proposed framework.

\subsection{Assessment Framework}

Deep learning-based deepfake detection methods currently rely heavily on the distribution of training data. Typically, these methods are evaluated using test datasets that are similar to the training sets. Some benchmarks additionally take real-world perturbations into consideration by randomly processing a subset of the test set and mixing it with other data. However, the results of these benchmarks are often stochastic and less reliable due to the absence of more realistic perturbations or a standard way of determining the proportion of corrupted data.


This section introduces the usage and principle of the proposed assessment framework in details. The deepfake detector is first trained on a target dataset without any modifications applied to it. Then, multiple copies of the test set are created. The influencing operations or distortions are applied to an entire copy of the test set respectively to avoid randomness in the results. The distorted data are then fed into the deepfake detector to calculate the performance metrics for each corresponding influencing factor. 
Furthermore, the computed metrics can be grouped according to the operation category to further analyze the impact of a specific processing operation.



\begin{table*}[t]
  \centering
  \caption{AUC (\%) scores of three selected deepfake detection methods on the distorted variants of the FFpp test set that are subject to different video processing operations. The notation C23 and C40 here refer to the two different compression rates using AVC/H.264 codec. Contrast refers to increasing the contrast by a fixed scale. Resolution refers to reducing video resolution by a specific scale.}
  \begin{adjustbox}{width=\textwidth}
    \begin{tabular}{ccccccccccccccc}
    \toprule
    \multirow{2}[4]{*}{Methods} & \multirow{2}[4]{*}{TrainSet} & \multicolumn{2}{c}{Compression} & \multicolumn{2}{c}{Brightness} & \multirow{2}[4]{*}{Contrast} & \multirow{2}[4]{*}{\shortstack{Gaussian \\ Noise}} & \multicolumn{2}{c}{Flipping} & \multicolumn{2}{c}{Resolution} & \multirow{2}[4]{*}{Grayscale} & \multirow{2}[4]{*}{\shortstack{Vintage \\ Filter}} & \multirow{2}[4]{*}{Average} \\
\cmidrule{3-6}\cmidrule{9-12}          &       & C23 & C40 & Increase & Decrease &       &       & Horizontal & Vertical & x2    & x4    &       &       &  \\
    \midrule
    \midrule
    CapsuleNet & \multirow{3}[2]{*}{FFpp-Raw} & 77.97 & 54.14 & 73.31 & 70.62 & 69.31 & 54.14 & 73.13 & 63.20 & 65.43 & 56.99 & 68.38 & 72.94 & 66.63 \\
    XceptionNet &       & 69.49 & 55.70 & 65.92 & 66.40 & 65.32 & 50.50 & 65.26 & 57.36 & 57.23 & 55.90 & 65.51 & 66.90 & 61.79 \\
    SBIs  &       & 90.43 & 76.27 & 86.38 & 86.47 & 85.94 & 71.52 & 85.98 & 79.28 & 76.35 & 63.62 & 86.27 & 86.54 & 81.25 \\
    \midrule
    CapsuleNet & \multirow{3}[2]{*}{FFpp-C23} & 95.61 & 66.03 & 93.27 & 92.31 & 91.55 & 53.50 & 91.98 & 71.49 & 80.28 & 67.56 & 87.43 & 88.86 & 81.66 \\
    XceptionNet &       & 98.34 & 70.71 & 97.07 & 96.65 & 96.34 & 51.04 & 96.20 & 66.82 & 83.42 & 72.03 & 93.17 & 94.99 & 84.73 \\
    SBIs  &       & 91.71 & 75.43 & 87.63 & 86.51 & 87.40 & 57.06 & 86.84 & 81.22 & 75.40 & 64.31 & 87.31 & 86.28 & 80.59 \\
    \midrule
    CapsuleNet & \multirow{2}[2]{*}{FFpp-C40}  & 82.64 & 78.33 & 80.22 & 80.77 & 79.30 & 52.78 & 78.64 & 61.53 & 76.88 & 71.91 & 78.41 & 75.82 & 74.77 \\
    XceptionNet &   & 83.25 & 80.69 & 80.85 & 82.83 & 80.65 & 51.74 & 81.39 & 55.70 & 80.62 & 74.99 & 71.30 & 78.43 & 75.20 \\
    \bottomrule
    \end{tabular}%
    \end{adjustbox}
  \label{tab:results-video}%
\end{table*}%

\section{Experimental Results}

\subsection{Implementation Details}

\subsubsection{Detection Methods}
Experiments have been conducted using the following three learning-based deepfake detection methods. 

\textbf{XceptionNet} \cite{roessler2019faceforensicspp} is a well-known CNN architecture and has been applied for detecting manipulated faces by functioning as a classification network.
The detection system is first pre-trained on ImageNet database \cite{5206848} and then re-trained on a deepfake detection dataset. It has become a popular baseline method in the FaceForenscis++ benchmark.

\textbf{CapsuleNet} is a deepfake detection method based on a combination of capsule network and conventional convolutional neural network. 
Nguyen et al. \cite{Nguyen2019UseOA} employed the capsule network in their deepfake detection pipeline and achieved the best performance at that time in the FaceForensics++ benchmark compared to other competing methods.

\textbf{SBIs} \cite{Shiohara_2022_CVPR} refers to a deepfake detection method based on synthetic data, called Self-Blended Images. The overall detection system is built on a pre-trained deep classification network, EfficientNet. During the training phase, the SBIs method generates hardly recognizable fake images that contain common face forgery traces to enforce the network to learn more general representation. It demonstrates state-of-the-art performance in cross-dataset evaluations.

\subsubsection{Datasets}
This paper adopts the FaceForensics++ dataset, denoted by FFpp, for extensive experimentation. It comprises 1000 pristine and 4000 manipulated video footage in three compression quality levels. 
In the experiment, the different quality data is used for training, denoted as \textit{FFpp-Raw}, \textit{FFpp-C23}, and \textit{FFpp-C40}.

\subsubsection{Training Details}
To train high-performing deepfake detectors, the following configurations have been selected. 
The XceptionNet and CapsuleNet are trained with Adam optimizer with $\beta_1=0.9$ and $\beta_2=0.999$ and a batch size of 64. The XceptionNet model is trained for 10 epochs using a learning rate of 0.001, while the CapsuleNet is trained for 25 epochs using a learning rate of 0.0005. For both methods, 100 frames are randomly sampled from each video for training purposes and 32 frames are extracted for validation and testing. 
The SBIs method has a different experimental setting. It is trained with SAM \cite{foret2020sharpness} optimizer for 100 epochs. The batch size and learning rate are set to 32 and 0.001 respectively. During the training phase, only authentic high-quality video is used and the corresponding fake samples are created by their proposed self-blending method. Only 8 frames per video are sampled for training while 32 frames are for validation and testing.

\subsection{Assessment Results}
This section reports a thorough assessment of three detection methods, i.e. CapsuleNet, XceptionNet, and SBIs, for identifying deepfakes under realistic situations using the framework described earlier. Table \ref{tab:results-video} summarizes the performance of the three deepfake detection methods.

As a result, different detection methods show variant characteristics under the assessment framework. Both the XcpetionNet and CapsuleNet heavily rely on the quality of the training set. When trained on very high- or low-quality data, both approaches exhibit poor performance when facing all kinds of processing operations.
However, their performance in our benchmark improves significantly and even surpasses the state-of-the-art SBIs method after training with slightly compressed training data. Under this setting, the changes in brightness, contrast, and color map bring limited impact, while heavy compression, low resolution, noise, and geometric transformation still remain to be significantly affecting factors.

On the contrary, the overall performance of the SBIs method declines by 0.6\% after training with compressed data, which implies that this method relies more on the difference between authentic training data and its fake counterparts. While XceptionNet and CapsuleNet suffer from vertical flipped video, it is notable that this operation shows a limited impact on the SBIs method. This is attributed to the fact that, unlike the other two detectors, the SBIs method leverages general traces of forgery rather than the global inconsistency on the face. On the other hand, none of the three methods can accurately classify deepfakes processed by heavy compression, resolution reduction, or video noise, which will be an open issue for both research and industrial deployment in deepfake detection.

\section{Conclusion}

Most of the current deepfake detection methods focus on achieving high performance on specific benchmarks. However, it has been shown that the assessment approaches employed in these benchmarks are less reliable and insightful. This paper analyzes in details numerous video processing operations and studies their impact on learning-based deepfake detection methods. Experiments have been performed on three popular detectors which bring insights into their future improvement.



\section*{Acknowledgment}

The authors acknowledge support from CHIST-ERA project XAIface (CHIST-ERA-19-XAI-011) with funding from
the Swiss National Science Foundation (SNSF) under grant number 20CH21 195532.

\newpage



\bibliographystyle{IEEEtran}
\bibliography{refs}

\begin{thebibliography}{10}
\providecommand{\url}[1]{#1}
\csname url@samestyle\endcsname
\providecommand{\newblock}{\relax}
\providecommand{\bibinfo}[2]{#2}
\providecommand{\BIBentrySTDinterwordspacing}{\spaceskip=0pt\relax}
\providecommand{\BIBentryALTinterwordstretchfactor}{4}
\providecommand{\BIBentryALTinterwordspacing}{\spaceskip=\fontdimen2\font plus
\BIBentryALTinterwordstretchfactor\fontdimen3\font minus
  \fontdimen4\font\relax}
\providecommand{\BIBforeignlanguage}[2]{{%
\expandafter\ifx\csname l@#1\endcsname\relax
\typeout{** WARNING: IEEEtran.bst: No hyphenation pattern has been}%
\typeout{** loaded for the language `#1'. Using the pattern for}%
\typeout{** the default language instead.}%
\else
\language=\csname l@#1\endcsname
\fi
#2}}
\providecommand{\BIBdecl}{\relax}
\BIBdecl

\bibitem{karras2017progressive}
T.~Karras, T.~Aila, S.~Laine, and J.~Lehtinen, ``Progressive growing of gans
  for improved quality, stability, and variation,'' \emph{arXiv preprint
  arXiv:1710.10196}, 2017.

\bibitem{karras2019style}
T.~Karras, S.~Laine, and T.~Aila, ``A style-based generator architecture for
  generative adversarial networks,'' in \emph{Proceedings of the IEEE/CVF
  conference on computer vision and pattern recognition}, 2019, pp. 4401--4410.

\bibitem{karras2020analyzing}
T.~Karras, S.~Laine, M.~Aittala, J.~Hellsten, J.~Lehtinen, and T.~Aila,
  ``Analyzing and improving the image quality of stylegan,'' in
  \emph{Proceedings of the IEEE/CVF conference on computer vision and pattern
  recognition}, 2020, pp. 8110--8119.

\bibitem{roessler2019faceforensicspp}
A.~R\"ossler, D.~Cozzolino, L.~Verdoliva, C.~Riess, J.~Thies, and
  M.~Nie{\ss}ner, ``Face{F}orensics++: Learning to detect manipulated facial
  images,'' in \emph{International Conference on Computer Vision (ICCV)}, 2019.

\bibitem{thies2019deferred}
J.~Thies, M.~Zollh{\"o}fer, and M.~Nie{\ss}ner, ``Deferred neural rendering:
  Image synthesis using neural textures,'' \emph{ACM Transactions on Graphics
  (TOG)}, vol.~38, no.~4, pp. 1--12, 2019.

\bibitem{nirkin2019fsgan}
Y.~Nirkin, Y.~Keller, and T.~Hassner, ``Fsgan: Subject agnostic face swapping
  and reenactment,'' in \emph{Proceedings of the IEEE/CVF international
  conference on computer vision}, 2019, pp. 7184--7193.

\bibitem{zakharov2019few}
E.~Zakharov, A.~Shysheya, E.~Burkov, and V.~Lempitsky, ``Few-shot adversarial
  learning of realistic neural talking head models,'' in \emph{Proceedings of
  the IEEE/CVF international conference on computer vision}, 2019, pp.
  9459--9468.

\bibitem{Nguyen2019UseOA}
H.~H. Nguyen, J.~Yamagishi, and I.~Echizen, ``Use of a capsule network to
  detect fake images and videos,'' \emph{ArXiv}, 2019.

\bibitem{Zhao2021MultiattentionalDD}
H.~Zhao, W.~Zhou, D.~Chen, T.~Wei, W.~Zhang, and N.~Yu, ``Multi-attentional
  deepfake detection,'' \emph{2021 IEEE/CVF Conference on Computer Vision and
  Pattern Recognition (CVPR)}, pp. 2185--2194, 2021.

\bibitem{liu2021spatial}
H.~Liu, X.~Li, W.~Zhou, Y.~Chen, Y.~He, H.~Xue, W.~Zhang, and N.~Yu,
  ``Spatial-phase shallow learning: rethinking face forgery detection in
  frequency domain,'' in \emph{Proceedings of the IEEE/CVF conference on
  computer vision and pattern recognition}, 2021, pp. 772--781.

\bibitem{qian2020thinking}
Y.~Qian, G.~Yin, L.~Sheng, Z.~Chen, and J.~Shao, ``Thinking in frequency: Face
  forgery detection by mining frequency-aware clues,'' in \emph{European
  conference on computer vision}.\hskip 1em plus 0.5em minus 0.4em\relax
  Springer, 2020, pp. 86--103.

\bibitem{li2021frequency}
J.~Li, H.~Xie, J.~Li, Z.~Wang, and Y.~Zhang, ``Frequency-aware discriminative
  feature learning supervised by single-center loss for face forgery
  detection,'' in \emph{Proceedings of the IEEE/CVF conference on computer
  vision and pattern recognition}, 2021, pp. 6458--6467.

\bibitem{luo2021generalizing}
Y.~Luo, Y.~Zhang, J.~Yan, and W.~Liu, ``Generalizing face forgery detection
  with high-frequency features,'' in \emph{Proceedings of the IEEE/CVF
  conference on computer vision and pattern recognition}, 2021, pp.
  16\,317--16\,326.

\bibitem{khodabakhsh2018fake}
A.~Khodabakhsh, R.~Ramachandra, K.~Raja, P.~Wasnik, and C.~Busch, ``Fake face
  detection methods: Can they be generalized?'' in \emph{2018 international
  conference of the biometrics special interest group (BIOSIG)}.\hskip 1em plus
  0.5em minus 0.4em\relax IEEE, 2018, pp. 1--6.

\bibitem{xuan2019generalization}
X.~Xuan, B.~Peng, W.~Wang, and J.~Dong, ``On the generalization of gan image
  forensics,'' in \emph{Chinese conference on biometric recognition}.\hskip 1em
  plus 0.5em minus 0.4em\relax Springer, 2019, pp. 134--141.

\bibitem{lu2022novel}
Y.~Lu and T.~Ebrahimi, ``A novel assessment framework for learning-based
  deepfake detectors in realistic conditions,'' in \emph{Applications of
  Digital Image Processing XLV}, vol. 12226.\hskip 1em plus 0.5em minus
  0.4em\relax SPIE, 2022, pp. 207--217.

\bibitem{Agarwal_2019_CVPR_Workshops}
S.~Agarwal, H.~Farid, Y.~Gu, M.~He, K.~Nagano, and H.~Li, ``Protecting world
  leaders against deep fakes,'' in \emph{Proceedings of the IEEE/CVF Conference
  on Computer Vision and Pattern Recognition (CVPR) Workshops}, June 2019.

\bibitem{Yang2019ExposingDF}
X.~Yang, Y.~Li, and S.~Lyu, ``Exposing deep fakes using inconsistent head
  poses,'' \emph{ICASSP 2019 - 2019 IEEE International Conference on Acoustics,
  Speech and Signal Processing (ICASSP)}, pp. 8261--8265, 2019.

\bibitem{9072088}
T.~Jung, S.~Kim, and K.~Kim, ``Deepvision: Deepfakes detection using human eye
  blinking pattern,'' \emph{IEEE Access}, vol.~8, pp. 83\,144--83\,154, 2020.

\bibitem{zhou_two-stream_2017}
P.~Zhou, X.~Han, V.~I. Morariu, and L.~S. Davis, ``Two-{Stream} {Neural}
  {Networks} for {Tampered} {Face} {Detection},'' in \emph{2017 {IEEE}
  {Conference} on {Computer} {Vision} and {Pattern} {Recognition} {Workshops}
  ({CVPRW})}, Jul. 2017, pp. 1831--1839, iSSN: 2160-7516.

\bibitem{guera2018deepfake}
D.~G{\"u}era and E.~J. Delp, ``Deepfake video detection using recurrent neural
  networks,'' in \emph{2018 15th IEEE international conference on advanced
  video and signal based surveillance (AVSS)}.\hskip 1em plus 0.5em minus
  0.4em\relax IEEE, 2018, pp. 1--6.

\bibitem{sabir2019recurrent}
E.~Sabir, J.~Cheng, A.~Jaiswal, W.~AbdAlmageed, I.~Masi, and P.~Natarajan,
  ``Recurrent convolutional strategies for face manipulation detection in
  videos,'' \emph{Interfaces (GUI)}, vol.~3, no.~1, pp. 80--87, 2019.

\bibitem{Montserrat_2020_CVPR_Workshops}
D.~M. Montserrat, H.~Hao, S.~K. Yarlagadda, S.~Baireddy, R.~Shao, J.~Horvath,
  E.~Bartusiak, J.~Yang, D.~Guera, F.~Zhu, and E.~J. Delp, ``Deepfakes
  detection with automatic face weighting,'' in \emph{Proceedings of the
  IEEE/CVF Conference on Computer Vision and Pattern Recognition (CVPR)
  Workshops}, June 2020.

\bibitem{Shiohara_2022_CVPR}
K.~Shiohara and T.~Yamasaki, ``Detecting deepfakes with self-blended images,''
  in \emph{Proceedings of the IEEE/CVF Conference on Computer Vision and
  Pattern Recognition (CVPR)}, June 2022, pp. 18\,720--18\,729.

\bibitem{zheng2021exploring}
Y.~Zheng, J.~Bao, D.~Chen, M.~Zeng, and F.~Wen, ``Exploring temporal coherence
  for more general video face forgery detection,'' in \emph{Proceedings of the
  IEEE/CVF International Conference on Computer Vision}, 2021, pp.
  15\,044--15\,054.

\bibitem{dang2020detection}
H.~Dang, F.~Liu, J.~Stehouwer, X.~Liu, and A.~K. Jain, ``On the detection of
  digital face manipulation,'' in \emph{Proceedings of the IEEE/CVF Conference
  on Computer Vision and Pattern recognition}, 2020, pp. 5781--5790.

\bibitem{Li2019OnLF}
P.~Li, L.~Prieto, D.~Mery, and P.~J. Flynn, ``On low-resolution face
  recognition in the wild: Comparisons and new techniques,'' \emph{IEEE
  Transactions on Information Forensics and Security}, vol.~14, pp. 2000--2012,
  2019.

\bibitem{yang2019exposing}
X.~Yang, Y.~Li, and S.~Lyu, ``Exposing deep fakes using inconsistent head
  poses,'' in \emph{ICASSP 2019-2019 IEEE International Conference on
  Acoustics, Speech and Signal Processing (ICASSP)}.\hskip 1em plus 0.5em minus
  0.4em\relax IEEE, 2019, pp. 8261--8265.

\bibitem{korshunov2018deepfakes}
P.~Korshunov and S.~Marcel, ``Deepfakes: a new threat to face recognition?
  assessment and detection,'' \emph{arXiv preprint arXiv:1812.08685}, 2018.

\bibitem{Celeb_DF_cvpr20}
Y.~Li, X.~Yang, P.~Sun, H.~Qi, and S.~Lyu, ``Celeb-df: A large-scale
  challenging dataset for deepfake forensics,'' in \emph{IEEE Conference on
  Computer Vision and Patten Recognition (CVPR)}, 2020.

\bibitem{jiang2020deeperforensics10}
L.~Jiang, R.~Li, W.~Wu, C.~Qian, and C.~C. Loy, ``Deeperforensics-1.0: A
  large-scale dataset for real-world face forgery detection,'' 2020.

\bibitem{DFDC2020}
B.~Dolhansky, J.~Bitton, B.~Pflaum, J.~Lu, R.~Howes, M.~Wang, and C.~C. Ferrer,
  ``The deepfake detection challenge dataset,'' 2020.

\bibitem{5206848}
J.~Deng, W.~Dong, R.~Socher, L.-J. Li, K.~Li, and L.~Fei-Fei, ``Imagenet: A
  large-scale hierarchical image database,'' in \emph{2009 IEEE Conference on
  Computer Vision and Pattern Recognition}, 2009, pp. 248--255.

\bibitem{foret2020sharpness}
P.~Foret, A.~Kleiner, H.~Mobahi, and B.~Neyshabur, ``Sharpness-aware
  minimization for efficiently improving generalization,'' \emph{arXiv preprint
  arXiv:2010.01412}, 2020.

\end{thebibliography}
%
%
%

\end{document}